\documentclass[lettersize,journal]{IEEEtran}
\usepackage[colorlinks=true, linkcolor=blue, citecolor=blue, urlcolor=blue]{hyperref}
\usepackage{lettrine}
\usepackage{amsmath,amsfonts}
\usepackage{algpseudocode}
\usepackage{algorithm}
\usepackage{array}
\usepackage[caption=false,font=normalsize,labelfont=sf,textfont=sf]{subfig}
\usepackage{textcomp}
\usepackage{stfloats}
\usepackage{url}
\usepackage{verbatim}
\usepackage{graphicx}
\usepackage{subcaption}
\usepackage{booktabs}
\usepackage{makecell}
\usepackage{multirow}

\usepackage{siunitx}
\usepackage{cite}
\usepackage{tikz}
\usepackage{tikz}
\usepackage{soul}
\usepackage{color}
\soulregister{\cite}7
\soulregister{\ref}7
\usetikzlibrary{decorations.pathreplacing}
\usepackage{hyperref} 

\begin{document}

\author{Peng Liu,
        Huibing Zeng,
        Yiqun~Zhang,~\IEEEmembership{Senior Member,~IEEE},
        Yang Yi,~\IEEEmembership{Member,~IEEE}
        and Jigang Wu

\thanks{Peng Liu, Huibing Zeng, Yiqun Zhang and Jigang Wu are with the School of Computer Science and Technology, Guangdong University of Technology, Guangzhou 510006, China (e-mail: liupeng@gdut.edu.cn, 3281165050@qq.com, yqzhang@gdut.edu.cn, asjgwucn@outlook.com).}
\thanks{Yang Yi is with the College of Information and Artificial Intelligence, Yangzhou University, Yangzhou 225000, China (e-mail: yiyang@yzu.edu.cn).}
\thanks{Yiqun Zhang is the corresponding author.}
}

\title{Resource-Efficient Pruning for Transformer via Low-Rank Importance Estimation}

\markboth{IEEE TRANSACTIONS ON EMERGING TOPICS IN COMPUTATIONAL INTELLIGENCE}%
{Liu \MakeLowercase{\textit{et al.}}:Resource-Efficient Pruning for Transformer via Low-Rank Importance Estimation}

\maketitle

\begin{abstract}
With the rapid development of large-scale pre-trained language models based on Transformer architectures, their high computational and memory costs have become a major obstacle to deployment, especially in resource-constrained environments. Traditional pruning methods typically depend on full gradient-based importance estimation, and they necessitate prior finetuning of the model to achieve satisfactory performance. This process often results in intolerable resource consumption.
This paper proposes REP-LIE, a new approach to enable resource-efficient pruning during the process of finetuning.
REP-LIE leverages the gradients of LoRA low-rank matrices to estimate the importance of weights without requiring full gradient computation. To address the inherent randomness in importance estimation, a stability score is introduced, serving as the basis for iterative pruning of unimportant model parameters.
The pruned model is further finetuned through lightweight updates, eliminating the need for full-parameter optimization in the process of finetuning.
Extensive experiments on both medium-scale encoder models and large-scale generative models (LLaMA-7B and Mistral-7B) demonstrate that REP-LIE still achieves competitive performance compared to existing approaches.
\end{abstract}


\section{Introduction}
\lettrine[lines=2, findent=2pt, nindent=0pt]{T}{ransformer-based} neural network architectures have achieved outstanding performance in a wide range of areas, including computer vision, natural language understanding, and multi-modal tasks\cite{vaswani2017attention}.  Notably, large-scale language models (LLMs)\cite{zhang2025trending} like BERT\cite{devlin2019bert}, RoBERTa\cite{liu2020roberta}, and GPT-3\cite{brown2020gpt3} have shown remarkable success in many downstream tasks\cite{zhang2026how}, including text classification\cite{wang2018glue}, reading comprehension, and summarization\cite{liu2019fine}. 
However, the exceptionally vast size of large models pose challenges for practical development. For instance, the GPT-3 model has 175 billion parameters, requiring approximately 350GB of storage. The sheer volume of parameters and the associated computational expenses necessitate the devices with exceedingly high memory and computational capabilities. Nevertheless, the stringent requirements for such advanced devices pose formidable barriers to the widespread practical implementation of these models. 
This challenge can be effectively addressed by reducing the number of parameters of deep neural networks, thus reducing the memory and computational costs for training and inference.

Recent studies have demonstrated that a certain level of parameter redundancy exists within the Transformer model and can be eliminated with negligible performance degradation\cite{voita2019analyzing,2020pruning}.
Motivated by these findings, many works have focused on reducing both parametric and structural redundancy in Transformer model\cite{lin2024awq,sot2023Dynamic,23vitpruning}. As a highly efficient model compression technique, network pruning has demonstrated its effectiveness. It significantly reduces the storage requirements and computational complexity of models through the removal of redundant components \cite{summary}.

Network pruning methods are commonly categorized as unstructured pruning and structured pruning, based on the granularity of removed components. Among these methods, both unstructured and structured pruning aim to estimate the importance of weights for the finetuning task and prune out the least important weights by the system-supplied sparsity ratio.
The unstructured pruning can yield competitive results. However, the deployment of acceleration is limited by irregular sparsity and hardware constraints\cite{han2016deep,2014DO,wang2024unstructured}. Therefore,
most of existing works focus on the structured pruning method, which reduces model size by removing structural units, such as self-attention heads, feed-forward layer channels, and layers\cite{liu2019collaborative,michel2019sixteen,kang2020accelerator}.
 
\begin{figure*}[htbp]
	\centering
	\includegraphics[width=0.9\textwidth, trim= 5 100 5 50 ,clip]{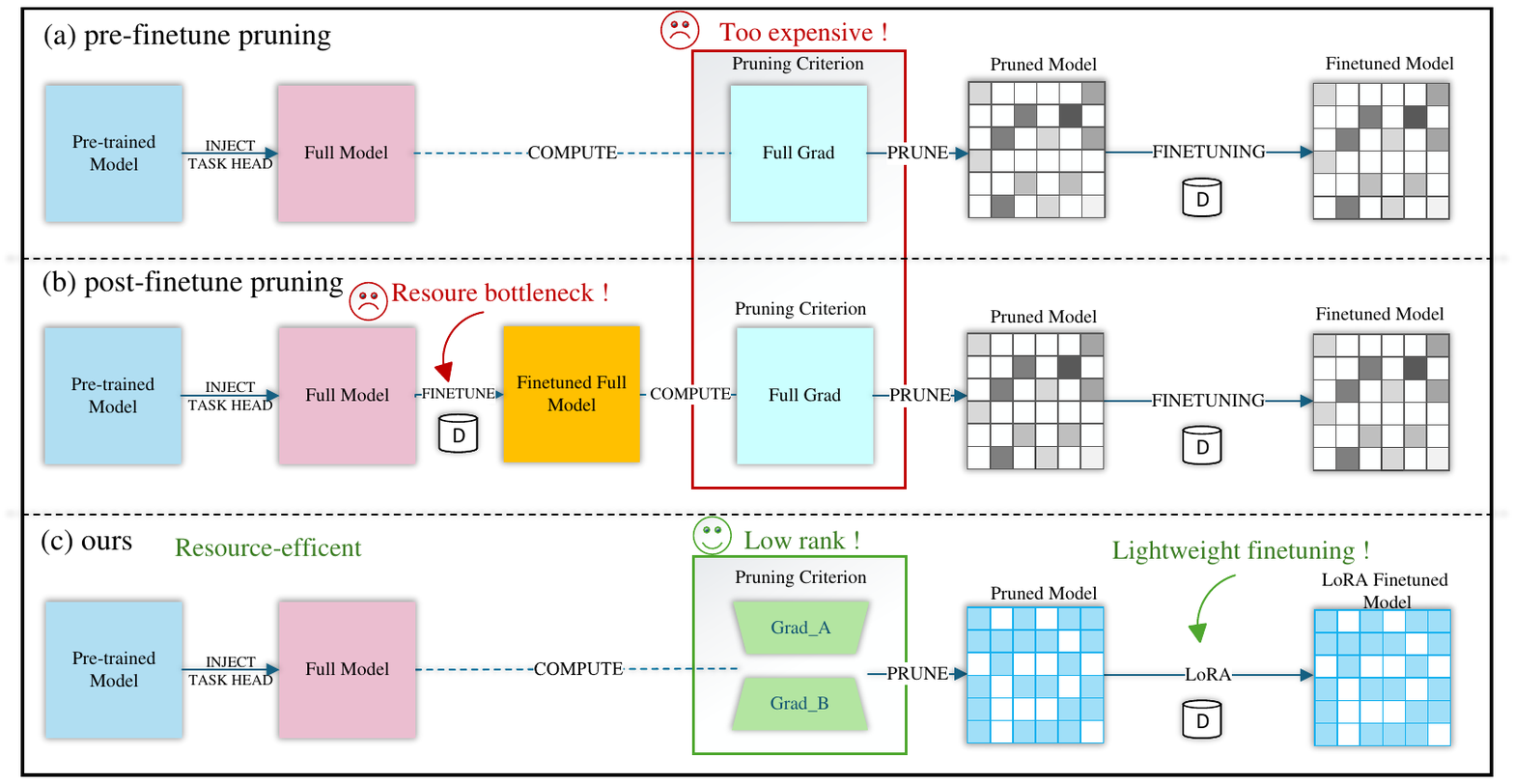}
	\caption{Comparison of pruning architecture. (a) pre-finetune pruning; (b) post-finetune pruning; (c) REP-LIE.}
	\label{fig:comparation}
\end{figure*}

The mainstream structured pruning strategies currently mainly fall into two categories: pre-finetune pruning and post-finetune pruning. The pre-finetune pruning strategy first prunes the full model and then finetunes the pruned model. In contrast, the post-finetune pruning strategy involves a two-stage finetuning process: first finetuning the full model, followed by finetuning the pruned model.
Fig.~\ref{fig:comparation} provides the workflows for the pre-finetune pruning and post-finetune pruning strategies.  
Both of them rely on gradient-based importance estimation and full-model finetuning. Considering the huge numbers of parameters in large language models, both the parameter importance estimation and full-model finetuning may impose substantial memory and computational overhead. Besides, the prolonged finetuning is required to restore performance for the structured pruning strategies \cite{frankle2018lottery}. These limitations underscore the urgent need for more effective pruning strategies.

Parameter-efficient finetuning methods such as LoRA (Low- Rank Adaptation) \cite{hu2022lora} provide a novel approach to address the aforementioned challenges. LoRA can significantly reduce the cost for finetuning, by freezing the original model weights and injecting a pair of trainable low-rank matrices into each layer of the model architecture. For example, LLM-Pruner\cite{ma2023llmpruner} significantly reduces finetuning cost by using LoRA modules to restore model performance after pruning. 
However, such methods still rely on the gradients to compute the importance of weights, which will lead to non-negligible pruning-induced overhead \cite{Xin2022grad,Wang2020grad}.
	
To address the trade-off between the model performance and the pruning overhead, we propose REP-LIE, a resource-efficient pruning framework for Transformer with low-rank importance estimation and lightweight finetuning, as shown in Fig.~\ref{fig:comparation}{\color{blue}(c)}. REP-LIE aims to the computational cost and the quantity of model parameters, while minimizing performance loss. 
In contrast to the traditional method that relies on gradient-based importance estimation, REP-LIE estimates the importance using only the gradients of low-rank matrices in LoRA. This effectively reduces the overhead during the process of pruning. Furthermore, distinct from conventional pre-finetune or post-finetune pruning strategies, REP-LIE capitalizes on the LoRA module's distinctive parameter-efficient adaptation mechanism. By updating low-rank metrics, REP-LIE establishes an effective pathway for restoring the performance of pruned models. This innovative strategy obviates the necessity of updating the entire parameter set of the model.
The main contributions are summarized as follows:
\begin{enumerate}
\item A new model compression method called REP-LIE is proposed. REP-LIE introduces a novel pruning criterion that evaluates the importance of weights solely using the gradients of low-rank matrices in LoRA. This fills the gap in existing methods that rely on full-model gradients computation for weight importance evaluation. Additionally, a concept of stability score is introduced for effective estimation of the weight importance. 
\item A lightweight finetuning strategy is designed to reduce the resource consumption, during the process of performance recovering for the pruned model. Unlike full-model fine-tuning adopted by most existing works, this strategy in REP-LIE only updates low-rank matrices to restore model performance, significantly cutting down GPU memory usage and training time during the recovery phase.

\item Experimental results demonstrate that REP-LIE still achieves competitive  performance across both medium-scale encoder models and large-scale generative models. 
This comprehensive evaluation 
demonstrates that REP-LIE achieves consistent performance advantages across all 
scales, confirming its scalability and robustness.
This cross-scale compatibility 
establishes REP-LIE as a sustainable compression paradigm. The resource-efficiency 
gains it provides remain applicable and effective regardless of future model 
scale evolution.
\end{enumerate}

The rest of this paper is organized as follows. Section~\ref{sec:relate} introduces the background of model pruning and the motivation of this work. Section \ref{sec:method} details the proposed structured pruning framework, including the criteria and the algorithm for pruning. Section \ref{sec:result} reports the experiment evaluations and Section \ref{sec:conclu} concludes the paper.

\section{Related Works}\label{sec:relate}
\subsection{Transformer Architecture}
This work focuses on the pruning for BERT model, which is one of the encoder-based Transformer models \cite{devlin2019bert}. Each encoder layer in BERT is composed of a multi-head attention (MHA) layer and a Feed-Forward Network (FFN) layer. The output of each layer can be formally presented as follows:

\begin{equation}
\resizebox{.90\hsize}{!}{$
	\mathrm{MHA}(x)=\sum_{h = 1}^{H}\mathrm{Attn}(W_q^{(l,h)},W_k^{(l,h)},W_v^{(l,h)},W_o^{(l,h)},x),
$}
\end{equation}

\begin{equation}
	X_{\mathrm{MHA}}=\mathrm{LayerNorm}(x+\mathrm{MHA}(x)),
\end{equation}

\begin{equation}
	\mathrm{FFN}(x)=\sigma(xW^{\mathrm{(1)}}+b^{\mathrm{(1)}})W^{\mathrm{(2)}}+b^{\mathrm{(2)}}.
\end{equation}

Specifically, the MHA layer consists of $H$ independently parameterized attention heads. Each head is associated with four projection matrices: a query matrix $W_Q \in \mathbb{R}^{d \times k}$, a key matrix $W_K \in \mathbb{R}^{d \times k}$, a value matrix $W_V \in \mathbb{R}^{d \times k}$, and an output projection matrix $W_O \in \mathbb{R}^{d \times k}$. Here, $d$ denotes the hidden dimension, $k$ is the output dimension of each self-attention head, which is calculated as $k=d/H$. The FFN layer typically consists of two projection matrices, where $\sigma$ denotes the activation function. 

\subsection{Pruning Criterion}
The estimation of parameter importance is wildly used to guide the process of pruning, by assessing the impact of individual parameter on the performance of neural networks.
Early works utilize the second-order Taylor expansion to estimate the impact of parameters on the loss function \cite{lecun1990obd}. However, it is expensive to compute second-order Taylor expansion in large-scale networks. To address the issue, some of studies have used the weight magnitude as an importance metric to prune the weights with small absolute values \cite{han2015learning, Gui2018Opt,Jae2021Layer,2024WORK}. Nevertheless, the small-magnitude weights may have a significant influence on the model performance, due to the inherent interactions of deep neural networks \cite{frankle2018lottery}.
To accurately capture the impact of pruning on model performance, Molchanov et al. have proposed an estimation method based on first-order Taylor expansion. They introduced gradient for weight importance assessment and validated its effectiveness in convolutional neural networks\cite{molchanov2019importance}. Inspired by this, many works have employed first-order gradients for importance estimation \cite{Xin2022grad,GRAD2025,Wang2020grad,zhang22ao}.
Additionally, to further approximate second-order information more effectively, some of studies have leveraged the Fisher information matrix as a proxy for Hessian\cite{KwonKMHKG2022fisher,Liu2021fisher,GLR2024}, while others have simplified gradient computation via clustering analysis \cite{gamanayake2020cluster}. Although these improvements have reduced the overhead of weight importance estimation, they still fail to overcome the high overhead caused by gradient computation and memory storage.
\subsection{Structured Pruning}
As a highly effective model compression approach, the structured pruning has been widely adopted in Transformer models in recent years.
To enhance structural reduction flexibility and task adaptability, EBERT uses an input-aware mechanism for sample-level dynamic pruning of attention heads and feed-forward network channels \cite{LiuLLC2021Ebert}. CoFi accelerates inference by integrating coarse-grained and fine-grained pruning with layer-by-layer distillation \cite{CoFiXia2022}.
ARC enhances model compression by combining fine-grained self-attention distillation with layer-wise random replacement training \cite{ARC2025}.
FLOP utilizes low-rank decomposition coupled with $L_{0}$ regularization and enhanced Lagrangian optimization to adaptively remove redundant elements from weight matrices\cite{2020-flop}. BMP innovates with semi-structured block-level pruning, compressing sub-matrices within the attention and feed-forward modules\cite{2021-bmp}.
\begin{figure}[t]                      
	\centering
	\includegraphics[width=\linewidth, trim = 0 100 0 120,clip]{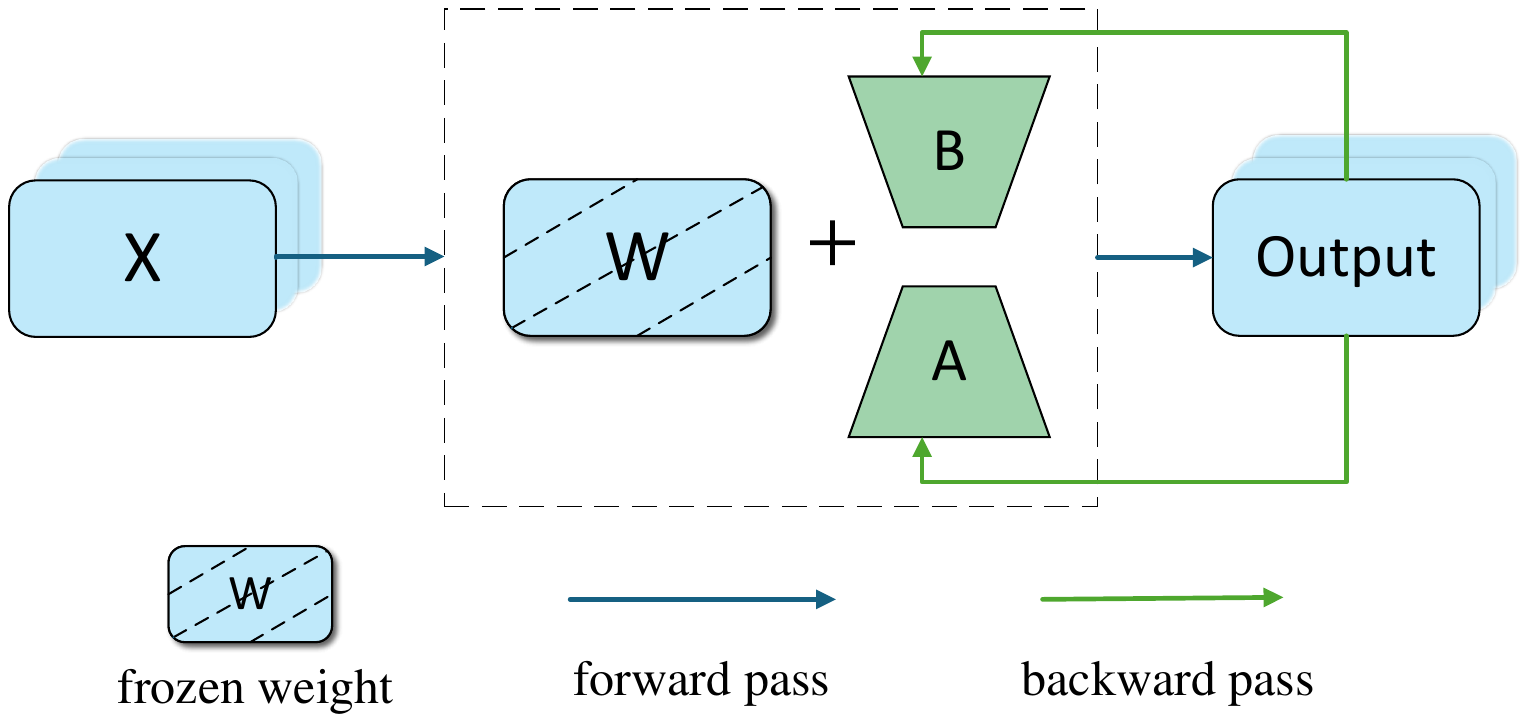}
	\caption{The schematics of LoRA on a single layer.}
	\label{fig:lora}	
\end{figure}
However, these methods still require full-model finetuning and rely on additional techniques like distillation to recover model performance.
To further reduce pruning cost, RECAP uses the second-order Taylor expansion-based weight importance estimation and then update a smaller subnetworks of the full model based on the Fisher information criterion\cite{Ilhan2024recap}. This approach saves memory overhead while maintaining model performance. However, it still relies on gradient-based importance estimation, which introduces additional overhead.
\subsection{Low-Rank Adaptation} 
\label{sec:lora}
As we pre-train larger models, it becomes less feasible to perform full fintuning due to their massive parameter size. LoRA emerges as a parameter-efficient approach, providing a promising solution to this issue, as depicted in Fig.~\ref{fig:lora}. LoRA freezes the original pre-trained weight matrix \(W\in\mathbb{R}^{d\times k}\) and introduces two low-rank matrices: \(A\in\mathbb{R}^{r\times k}\) and \(B\in\mathbb{R}^{d\times r}\), where \(r\ll\min(d,k)\). The adapted forward computation is expressed as:
\begin{equation}
	\textit{output} = xW_0+xBA=x(W_0+\underbrace{BA}_{\Delta W})
\end{equation}
where \(x\in\mathbb{R}^{n\times d}\) represents the input features. LoRA reduces the trainable parameters complexity from \(\mathcal{O}(dk)\) to \(\mathcal{O}((d + k)r)\), achieving significant efficiency improvements. In this reparameterization \(W = W_0 + BA\), \(A\) is initialized with a Gaussian distribution, and \(B\) with zeros to ensure the initial output remains unchanged (i.e., \(\Delta W = 0\)).

Studies show that using a low rank (e.g., \(r = 8\)) achieves performance close to full finetuning, with less than 0.1\% increase in parameters~\cite{benedek2024prilora}.  
Hu et al.\ apply LoRA to the \(W_Q\) and \(W_V\) matrices in MHA~\cite{hu2022lora}, while other works extend it to feed-forward layers for further gains~\cite{He2022lora}.
 
\begin{figure*}[htbp]
	\centering
	\includegraphics[width=0.9\textwidth, trim = 0 50 0 0,clip]{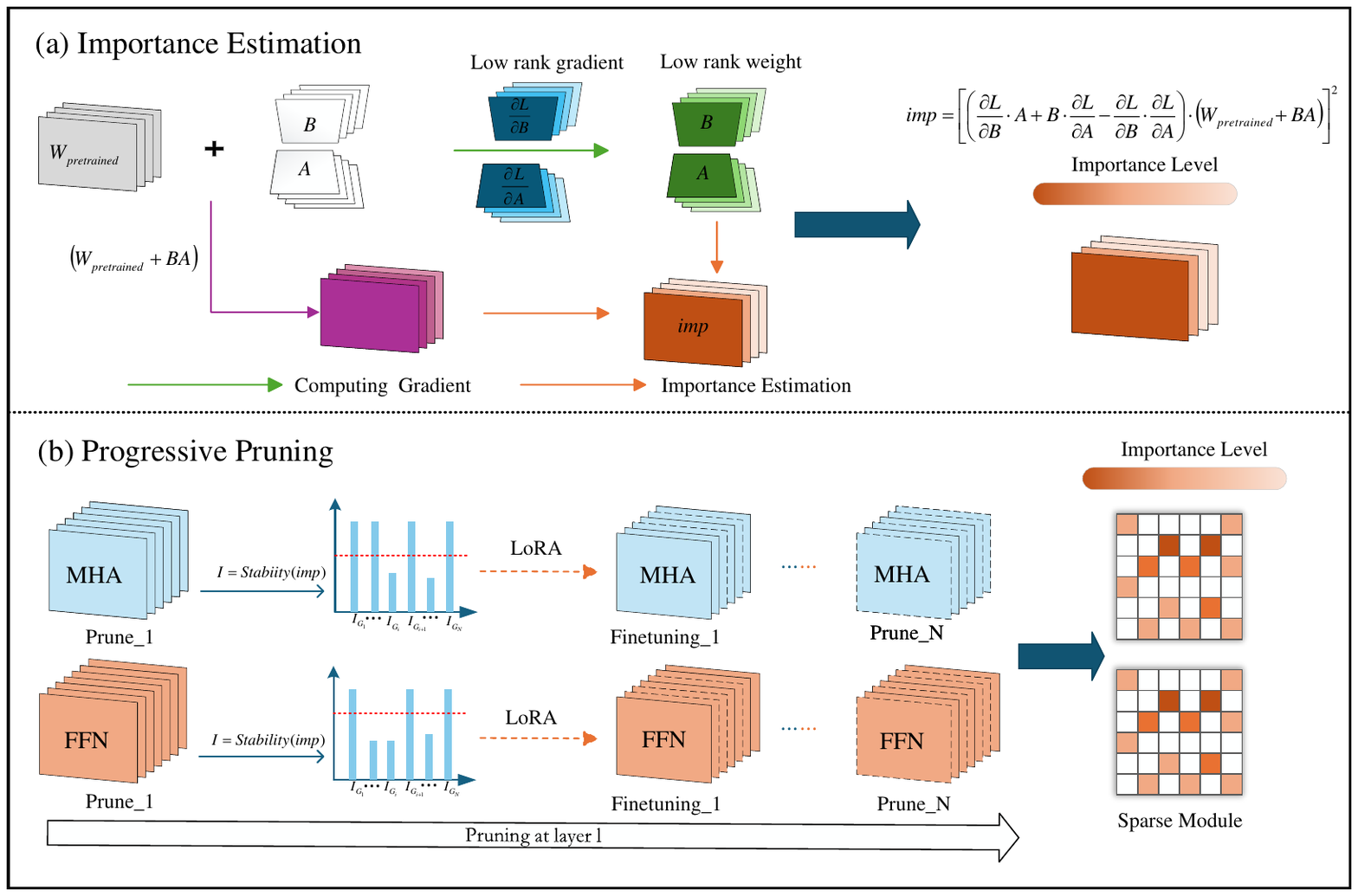}
	\caption{Overall framework diagram of REP-LIE. (a) Employing the gradients of LoRA's low-rank matrices to estimate importance; (b) Stability-Aware progressive structured pruning.}
	\label{fig:method}
\end{figure*}

\section{Proposed Method}\label{sec:method}
 In this section, we propose REP-LIE, a structured pruning framework that leverages low-rank gradients for importance estimation and employs a lightweight finetuning strategy during the progressive pruning process. REP-LIE aims to achieve efficient model compression and performance recovery without incurring substantial computational or memory overhead. The overview of REP-LIE is illustrated in Fig.~\ref{fig:method}. Firstly, gradients from low-rank matrices are utilized to evaluate the importance of weights, as shown in Fig.~\ref{fig:method}{\color{blue}(a)}. Then, a stability score is introduced to mitigate randomness in importance estimation, as shown in Fig.~ \ref{fig:method}{\color{blue}(b)}. Unimportant model parameters are iteratively pruned based on these stability scores. After each pruning round, model performance is restored by updating only the LoRA parameters.

\subsection{Grouping Model Weights}
REP-LIE employs structured pruning to enhance hardware efficiency and enable the practical employment of the model to hardware. For Transformer encoders, especially those with BERT-family architectures, we select the parameter in MHA and FFN as candidates for pruning. In the MHA layer, each attention head, which consists of the corresponding query, key, value, and output projection matrices, is regarded as a pruning unit. In the FFN layer, each hidden unit (i.e., the neuron) serves as the unit to be pruned, as shown in Fig.~\ref{fig:method}{\color{blue}(b)}. After grouping the model weights, we assign a binary mask variable to each group of modules. The mask determines if the corresponding group should be pruned.
The multi-head masked attention and the masked FFN computation are given by:
\begin{align}
\text{MHA}({{x}};\mathbf{m}_{l}^{\text{MHA}}) 
&= \sum_{i = 1}^{H} m_{l,i}^{\text{MHA}} \circ \text{Attn}_i({{x}}) \label{eq:mha}, 
\end{align}
\begin{align}
	\resizebox{.90\hsize}{!}{$
		\text{FFN}({{x}};\mathbf{m}_{l}^{\text{FFN}})
		= \left( \sum_{i = 1}^{N} m_{l,i}^{\text{FFN}} \circ \mathbf{W}_{:,i}^{(2)} \, 
		\sigma\left( \mathbf{W}_{:,i}^{(1)} {{x}} + \mathbf{b}^{(1)} \right) \right)
		+ \mathbf{b}^{(2)}
		$} \label{eq:ffn}
\end{align}
where \(\mathbf{m}_{l}^{\text{MHA}} \in \mathbb{R}^H\) and \(\mathbf {m}_{l}^{\text{FFN}} \in \mathbb{R}^N\) denote the mask variables for the MHA and FFN in the \(l\)-th layer, respectively. Here, \({m}_{l,i}^{\text{MHA}}\) and \(m_{l,i}^{\text{FFN}}\) represent their \(i\)-th unit.
\subsection{Low-Rank Pruning Criterion}\label{pruing method}
In this section, we propose a lightweight importance estimation criterion derived from the low-rank gradients in LoRA. The proposed criterion innovatively integrates Taylor expansion for computational efficiency with low-rank gradients, enabling effective parameter importance assessment with negligible computational overhead. To enable a resource-efficient pruning criterion, we begin with a loss-based parameter importance estimation, then introduce the classical first-order Taylor approximation to reduce computational cost, and ultimately develop a novel low-rank adaptation-based formulation that enables highly efficient pruning with negligible overhead.

\subsubsection{Importance Estimation via Loss Function}
Building upon the proven efficacy of the Taylor expansion-based weight importance estimation method~\cite{Molchanov17}, we assume that the loss function can be locally approximated to reflect the effect of pruning individual parameters. The importance of a weight $W_{i,j} \in W_{0}$ can be quantified by measuring the impact of its removal on the loss function:
\begin{equation}
I_{i,j} = \left[ \mathcal{L}(x,y,\mathbf{W}) - \mathcal{L}\left(x,y,\mathbf{W}\,\Big|\,W_{i,j} = 0 \right) \right]^2.
\label{eq:loss}
\end{equation}

However, it is infeasible to compute the exact loss difference for each parameter. To address this issue, we adopt the first-order Taylor approximation.
\subsubsection{First-Order Taylor Approximation}
Based on first-order Taylor expansion, the importance $I_{i,j}$ can be approximated by the product of the weight and its gradient:
\begin{equation}
\mathcal{I}_{i,j} \approx \left| W_{i,j} \cdot \frac{\partial \mathcal{L}}{\partial W_{i,j}} \right|^2.
\end{equation}

This approximation not only reflects the impact of weights on model performance but also accounts for their magnitudes, thereby offering an reasonable estimation of parameter importance. However, the first-order term $\frac{\partial \mathcal{L}}{\partial W_{i,j}}$ still requires gradients of the pre-trained weights, which will incur non-negligible memory and computational overhead.
\subsubsection{Low-Rank Adaptation Approximation}
To further enhance the efficiency of importance estimation, a lightweight strategy is proposed that incorporates LoRA into the estimation process, thereby effectively minimizing additional computational overhead. Specifically, for the weights to be pruned, we assume a low-rank compensation term such that $-(\mathbf{B}\mathbf{A})_{i,j} = W_{i,j}$. Under this assumption, the loss difference in Eq.~\ref{eq:loss} can be reformulated as:
\begin{equation}
\hat{\mathcal{I}}_{i,j} = \left[ \mathcal{L}(x, y, \mathbf{W}) - \mathcal{L}\left(x, y, \mathbf{W} \,\Big|\, W_{i,j} = - (\mathbf{B}\mathbf{A})_{i,j} \right) \right]^2.
\end{equation}

Leveraging the first-order Taylor expansion, the approximated importance is calculated as:

\begin{equation}
\hat{\mathcal{I}}_{i,j} = \left[ \frac{\partial \mathcal{L}}{\partial (\mathbf{B}\mathbf{A})_{i,j}} \cdot \left( W_{i,j} + (\mathbf{B}\mathbf{A})_{i,j} \right) \right]^2.
\label{eq:10}
\end{equation}

However, as shown in Eq. \ref{eq:10}, preserving $\frac{\partial \mathcal{L}}{\partial (\mathbf{B}\mathbf{A})_{i,j}}$ still entails the same level of complexity as $\frac{\partial \mathcal{L}}{\partial \mathbf{W}_{i,j}}$ since BA shares the same shape of $\textbf{W}$.
Notably, the gradient $\frac{\partial \mathcal{L}}{\partial (\mathbf{B}\mathbf{A})_{i,j}}$ corresponds to the update of $(\mathbf{B}\mathbf{A})_{i,j}$ in the optimization process, and can be approximated by:

\begin{equation}
\frac{\partial \mathcal{L}}{\partial (\mathbf{B}\mathbf{A})_{i,j}} \propto (\mathbf{B}\mathbf{A})_{i,j}^{(t-1)} - (\mathbf{B}\mathbf{A})_{i,j}^{(t)}.
\end{equation}

To obtain a more precise form, the gradient $\frac{\partial \mathcal{L}}{\partial (\mathbf{B}\mathbf{A})_{i,j}}$ is expanded using the chain rule:

\begin{align}
	\resizebox{.89\hsize}{!}{$
	\hat{\mathcal{I}}_{i,j} = \Bigg[ 
    \left( 
        \frac{\partial \mathcal{L}}{\partial \mathbf{B}_{i,:}} \mathbf{A}_{:,j} 
        + \mathbf{B}_{i,:} \frac{\partial \mathcal{L}}{\partial \mathbf{A}_{:,j}} 
        - \frac{\partial \mathcal{L}}{\partial \mathbf{B}_{i,:}} \frac{\partial \mathcal{L}}{\partial \mathbf{A}_{:,j}} 
    \right) 
    \cdot \left( 
        W_{i,j} + (\mathbf{B}\mathbf{A})_{i,j} 
    \right) \Bigg]^2.
    	$}
    \label{eq:final}
\end{align}

The importance of each pruning unit is computed via Eq. \ref{eq:final}, as illustrated in Fig.~\ref{fig:method}{\color{blue}(a)}. This formulation leverages only the LoRA parameters ${\textit{A}}$ and ${\textit{B}}$. This makes it highly efficient for importance estimation with minimal memory and computational overhead. 

	\textbf{\textit{Remark 1 (Compatibility of REP-LIE):}} 
	REP-LIE is
	principally applicable to any Transformer model that can be equipped with
	LoRA modules. REP-LIE inherits
	the broad compatibility of LoRA itself: LoRA has been successfully applied
	to encoder-only models (BERT, RoBERTa, DeBERTa), decoder-only models
	(LLaMA, Mistral, GPT), and encoder-decoder models (T5, BART, NMT) across
	diverse tasks\cite{hu2022lora, 2024summary}. For models without native LoRA support, LoRA modules can be
	injected into all linear projection layers, including $W_q$, $W_k$, $W_v$ and $W_o$ in
	multi-head attention, as well as the up/down projections in FFN. This is achieved by replacing the original weight matrix
	$W \in \mathbb{R}^{d \times k}$ with $W_0 + BA$, where $A \in \mathbb{R}^{r \times k}$
	and $B \in \mathbb{R}^{d \times r}$. The additional trainable parameters scale
	with rank $r$: for BERT-scale models (109M parameters), rank $r=8$ introduces
	approximately 1--2\% overhead, which is negligible compared to the full
	fine-tuning cost and scales linearly with $r$. Crucially, this injection
	requires no architectural modification to the original model.

\subsection{Stabilization Mechanism}
During the iterative computations for the importance of pruning units, inconsistent results may arise due to two primary factors: the inherent randomness during the process of updating parameters and the varying sequence of input data batches. Although this randomness has a negligible effect on the importance scores of highly significant units, it can substantially influence those of less significant ones. To tackle this problem, we propose a stability score to evaluate the impact of randomness on each pruning unit. The calculation formula is presented as:
\begin{equation}
	\label{eq:stable}
\text{\textit{Stability}}_i = \frac{ \sum_{j=1}^{n} \left( \mathrm{Rank}_j(\mathrm{\hat{\mathcal{I}}}_i) - \mathrm{Rank_{\text{mean}}}(\mathrm{\hat{\mathcal{I}}}_i) \right) }{n}
\end{equation}
where $n$ denotes the number of iterations for importance calculation, $\text{Rank}_j(\mathrm{\hat{\mathcal{I}}}_i)$  denotes the ranking of the $i$-th pruning units among all pruning units in the $j$-th computation, and $\text{Rank}_{\text{mean}}(\mathrm{\hat{\mathcal{I}}}_i)$ stands for the average ranking of the $i$-th pruning units. The higher $\text{\textit{Stability}}_i$ values indicate more unstable units, which will reserve more instability factors in the model performance. In general, the higher $\text{\textit{Stability}}_i$ indicates a lower level of importance. 

\subsection{Adaptive Pruning under Resource Constraints}\label{progressive pruning}
To further enhance the deployment efficiency of Transformer models, we introduce resource-aware constraints and dynamic sparsity scheduling into the pruning process. Specifically, the pruning for models are guided by incorporating the computational budget constraints (e.g., FLOPs) and the sparsity constraints. 

The pruning problem is formulated as the following constrained optimization:
\begin{align}
	\min_{\mathbf{m}} \quad & \mathcal{L}_{\text{task}}(\mathbf{m}; \mathbf{W}) \notag\ \\
	\text{s.t.} \quad & \sum{l = 1}^{L} \left( \sum_{h = 1}^{H} m_h^{(l)} C_h^{(l)} + \sum_{f = 1}^{d_{\text{ff}}} m_f^{(l)} C_f^{(l)} \right) \leq C
\end{align}
where \(\mathcal{L}_{\text{task}}\) is the task loss under pruning mask \({m}\) and weights \({W}\), \({m}\) is a binary mask controlling attention heads and feed-forward units, \(C_h^{(l)}\) and \(C_f^{(l)}\) are the resource cost of retaining components, and \(C\) is the total resource budget.

The dynamic sparsity scheduling function, denoted as ${Scheduling}(t_i, t_f, T, v_i, v_f)$, is crucial for controlling the sparsity during training. It consists of three phases:

\begin{align} 
	\text{V}(t) = 
	\begin{cases}
		v_i, & t \in [0, t_i) \\
		v_f + (v_i - v_f)\left( \frac{T - t_f - t}{T - t_f - t_i} \right)^3, & t \in [t_i, T - t_f) \\
		v_f, & \text{otherwise}.
	\end{cases}
\end{align}
\begin{algorithm}[t]
	\caption{Adaptive Pruning with Dynamic Sparsity Scheduling}
	\label{alg:adaptive_pruning}
	\begin{algorithmic}[1]
		\Require Stability metric $I$, initial rate $v_i$, final rate $v_f$, total steps $T$, model
		\Ensure Optimal mask $m^* $
		\State Initialize head mask $m^h \gets \mathbf{1}$, FFN mask $m^f \gets \mathbf{1}$
		\State $\text{current\_rate} \gets 0$, $\text{target\_rate} \gets 0$
		\For{$step = 0$ to $T-1$}
		\State \parbox[t]{\dimexpr\linewidth-\algorithmicindent}{%
			\footnotesize
			$target\_rate\gets target\_rate+\mathit{Scheduling}(t_i, t_f, T, v_i, v_f)$
		}
		
		\State $N_h \gets \lfloor (1 - target\_rate) \cdot L_H \rfloor$
		\State $N_f \gets \lfloor (1 - target\_rate) \cdot L_F \rfloor$
		\State $HI \gets$ indices of $N_h$ least important heads from $I$
		\State $FI \gets$ indices of $N_f$ least important filters from $I$
		\State $S[step] \gets \text{model}(m^h, m^f)$ \Comment{Evaluate performance}
		\State $R[step] \gets (HI, FI)$ \Comment{Store current pruning plan}
		\If{$current\_rate < target\_rate$}
		\State Update $m^h$ by setting $m^h[HI] \gets 0$
		\State Update $m^f$ by setting $m^f[FI] \gets 0$
		\State $current\_rate \gets 1 - \frac{\sum m^h + \sum m^f}{L_H + L_F}$
		\EndIf
		\EndFor
		\State $n^* \gets \arg\max_n S[n]$
		\State $(HI^*, FI^*) \gets R[n^*]$
		\State Prune heads $m^h[HI^*] \gets 0$, filters $m^f[FI^*] \gets 0$
		\State \Return $m^* = (m^h, m^f)$
	\end{algorithmic}
\end{algorithm}

Algorithm \ref{alg:adaptive_pruning} presents the progressive pruning approach with dynamic sparsity scheduling. It generates the optimal binary mask \(m^*\) by iteratively adjusting the pruning ratio. In each iteration, the algorithm calculates the \(target\_rate\), \(N_h\) and \(N_f\). Then it selects the least important heads and units based on weight Stability \(I\). Next, the algorithm updates \((m^h, m^f)\) and \(current\_rate\), while simultaneously recording \(S[step]\) and \(R[step]\). Once all iterations have been completed, the step with the highest accuracy is chosen to extract \(HI^*\) and \(FI^*\), outputting the optimal \(m^*=(m^h, m^f)\) for the target sparsity.

\section{Experimental Evaluation}\label{sec:result}
To comprehensively evaluate REP-LIE, we conduct experiments across three representative model families and address six key research questions regarding its effectiveness, efficiency, scalability, and design choices.
\subsection{Experimental Settings}
Our experiments are structured to address the following key questions, with corresponding results reported in the indicated subsections:

	 \textbf{Q1 (Medium-scale performance)}: How does REP-LIE perform on encoder models? To ensure a fair comparison, data augmentation is excluded from TinyBERT. REP-LIE is compared with other compression methods in terms of performance across various GLUE tasks, and statistical significance is analyzed using a one-tailed Wilcoxon signed-rank test. The results show that all compression methods exhibit an inevitable performance drop when the model size is reduced, while REP-LIE is significantly superior to all other methods. Results are reported in Section~\ref{sec:50prune}.
	
	 \textbf{Q2 (Large-scale scalability)}: Does REP-LIE maintain competitive performance on large-scale generative models under varying compression levels? We evaluate both language modeling capability (perplexity on WikiText-2) and zero-shot commonsense reasoning across seven diverse tasks. The results demonstrate whether REP-LIE maintains competitive performance as model scale and architectural complexity increase, and how performance varies across sparsity levels and task types, as shown in Section \ref{sec:lager}.
	
	 \textbf{Q3 (High-compression robustness)}: How robust is REP-LIE under aggressive compression ratios? To investigate this, the performance of REP-LIE is evaluated under high sparsity levels and varying FLOPs reduction ratios, with results reported in Section \ref{sec:mehigh}. REP-LIE maintains stable performance across GLUE tasks at both high sparsity ratios and varying FLOPs reduction ratios, significantly outperforming mainstream compression methods such as CoFi and EBERT.
	
	 \textbf{Q4 (Memory efficiency)}: How does REP-LIE compare to other pruning frameworks in terms of memory efficiency? We measure the GPU memory footprint of REP-LIE and compare it with that of Head-FT, Pre-FT, Post-FT, and RECAP under a 50\% sparsity rate. Experimental results in Table \ref{tab:nlu_performance} show that REP-LIE achieves the lowest memory footprint (476MB), while maintaining competitive performance, as shown in Section \ref{sec:memory}.
	
	 \textbf{Q5 (Pruning and inference efficiency)}: What are the computational overhead and inference speedup of REP-LIE across different model scales? We report pruning runtime and peak GPU memory usage across BERT-base, LLaMA-7B, and Mistral-7B. For LLaMA-7B, we further measure actual inference latency, throughput, and memory consumption on real hardware. The results confirm that REP-LIE directly translates parameter reduction into tangible deployment gains, as shown in Section \ref{sec:efficiency}.
	
	 \textbf{Q6 (Criterion effectiveness)}: How effective is low-rank gradient-based importance estimation compared to alternative pruning criteria? We compare the model performance and resource usage corresponding to three different pruning criteria (weight magnitude, gradient-based, and LoRA-based), and visualize the model structures after pruning with these three criteria. The LoRA-based method outperforms the other two methods in terms of the average performance and memory efficiency. Additionally, we quantify the approximation error of low-rank gradients via Spearman correlation analysis across varying ranks, as shown in Section \ref{sec:criterion}.

\begin{table*}[!t]
	\small 
	\centering
	\caption{Model compression results with 50\% sparsity.\label{tab:50prune}}
	\resizebox{0.90\hsize}{!}{
	\begin{tabular}{l r ccccccc}
		\toprule
		\multirow{2}{*}{Method} & \multirow{2}{*}{\#Param} & \multicolumn{6}{c}{GLUE benchmark} & \multirow{2}{*}{Wilcoxon Sig. vs REP-LIE} \\
		\cmidrule(lr){3-8}
		& & QQP & QNLI & SST-2 & CoLA & MRPC & RTE \\
		\midrule
		BERT-base  & 109M & 91.50 & 91.40 & 93.23 & 58.90 & 88.48 & 66.80 & $p > 0.05$ \\
		\midrule
		CoFi       & 67M  & 90.70 & 89.10 & 91.20 & 52.40 & 87.30 & 64.60 & pass ($p = 0.0156$) \\
		PGB        & 67M  & 91.10 & 90.21 & 92.27 & 55.01 & 84.26 & 64.25 & pass ($p = 0.0156$) \\
		DynaBERT   & 67M  & 90.68 & 88.23 & 91.47 & 51.18 & 77.90 & 63.21 & pass ($p = 0.0156$) \\
		EBERT      & 67M  & 90.49 & 89.74 & 90.68 & N.A.  & 72.08 & 52.75 & pass ($p = 0.0312$) \\
		TinyBERT   & 67M  & 90.80 & 90.60 & 91.30 & 43.70 & 87.20 & 59.60 & pass ($p = 0.0156$) \\
		RECAP      & 67M  & 91.29 & 90.74 & 92.20 & 56.56 & 86.56 & 65.98 & pass ($p = 0.0156$) \\
		\textbf{REP-LIE(ours)}       & 67M  & \textbf{91.33} & \textbf{90.96} & \textbf{92.33} & \textbf{57.23} & \textbf{87.90} & \textbf{66.45} & -- \\
		\bottomrule
	\end{tabular}}
\end{table*}
\begin{table*}[!t]
	\centering
	\caption{PPL \& Commonsense Reasoning zero-shot performance of the pruned LLaMA-7B. The PPL is evaluated on Wikitext2 and the Avg. denotes mean accuracy across BoolQ--OBQA.}
	\label{tab:results}
	\small
    \resizebox{0.95\hsize}{!}{
	\begin{tabular}{c l c c c c c c c c c c c}
		\toprule
		Sparsity & Method & \#Params & PPL$\downarrow$ & BoolQ & PIQA & HellaS & WinoG & ARC-e & ARC-c & OBQA & Avg. & \makecell{Wilcoxon Sig.\\ vs REP-LIE}  \\
		\midrule
		\multirow{1}{*}{0\%}
		& LLaMA-7B & 6.7B & 12.63 & 75.08 & 79.16 & 76.20 & 70.00 & 72.89 & 44.88 & 44.40 & 66.09 & -- \\
		\midrule
		\multirow{4}{*}{20\%}
		& LLM-Pruner & \multirow{4}{*}{5.4B} & 18.01 & 66.76 & 78.45 & 71.44 & 63.77 & 66.41 & 39.85 & 43.80 & 61.50 &pass ($p = 0.0078$)  \\
		& Compresso  & & --    & \textbf{79.08} & 75.46 & 53.44 & 67.80 & 68.64 & 37.97 & 34.20 & 59.51 &pass ($p = 0.0156$)\\
		& SlimGPT    & & 16.68 & 74.59 & 78.94 & 74.40 & 68.43 & 70.50 & 43.26 & 45.40 & 65.07 &pass ($p = 0.0156$)\\
		& \textbf{REP-LIE(ours)}    &  & \textbf{16.32} & 78.78 & \textbf{79.91} & \textbf{75.12} & \textbf{69.14} & \textbf{75.93} & \textbf{43.17} & \textbf{45.52} & \textbf{66.80} &--\\
		\midrule
		\multirow{4}{*}{25\%}
		& LLM-Pruner & \multirow{4}{*}{5.0B} & 20.57 & 62.81 & 76.93 & 69.21 & 60.46 & 63.34 & 38.14 & 39.80 & 58.67 &pass ($p = 0.0078$)\\
		& Compresso  & & --    & 73.55 & 73.07 & 49.16 & 64.80 & 66.20 & 37.20 & 29.80 & 56.25 &pass ($p = 0.0078$)\\
		& SlimGPT    & & 18.45 & 73.46 & 77.42 & 72.07 & 65.51 & 67.17 & 41.13 & 40.40 & 62.45 &pass ($p = 0.0078$)\\
		& \textbf{REP-LIE(ours)}    &  & \textbf{18.03} & \textbf{78.71} & \textbf{77.91} & \textbf{73.47} & \textbf{69.30} & \textbf{75.72} & \textbf{43.09} & \textbf{43.25} & \textbf{65.96} &--\\
		\midrule
		\multirow{4}{*}{33\%}
		& LLM-Pruner & \multirow{4}{*}{4.5B} & 24.50 & 62.02 & 74.92 & 64.41 & 61.80 & 53.79 & 32.00 & 38.80 & 55.39 &pass ($p = 0.0078$)\\
		& Compresso  &  & --    & 68.69 & 72.85 & 47.18 & 63.38 & 65.99 & 35.07 & 29.00 & 54.59 &pass ($p = 0.0078$)\\
		& SlimGPT    &  & 22.43 & 71.53 & 76.66 & 70.55 & 66.06 & 64.35 & 39.33 & 41.40 & 61.41 &pass ($p = 0.0078$)\\
		& \textbf{REP-LIE(ours)}    &  & \textbf{20.39} & \textbf{74.23} & \textbf{77.41} & \textbf{71.28} & \textbf{68.05} & \textbf{70.88} & \textbf{40.52} & \textbf{42.31} & \textbf{63.52} \\
		\midrule
		\multirow{3}{*}{50\%}
		& LLM-Pruner & \multirow{3}{*}{3.4B} & 40.64 & 60.21 & 68.88 & 47.86 & 54.62 & 43.94 & 27.73 & 35.20 & 48.35 &pass ($p = 0.0078$)\\
		& SlimGPT    & & 31.07 & 65.11 & 71.60 & 59.94 & 59.27 & 53.37 & 31.83 & 35.20 & 53.76 &pass ($p = 0.0078$)\\
		& \textbf{REP-LIE(ours)}    &  & \textbf{28.46} & \textbf{70.58} & \textbf{73.26} & \textbf{60.23} & \textbf{60.98} & \textbf{58.52} & \textbf{32.40} & \textbf{40.02} & \textbf{56.57} &--\\
		\bottomrule
	\end{tabular}}
\end{table*}
\begin{figure*}[h]
	\centering
	\includegraphics[scale=0.4]{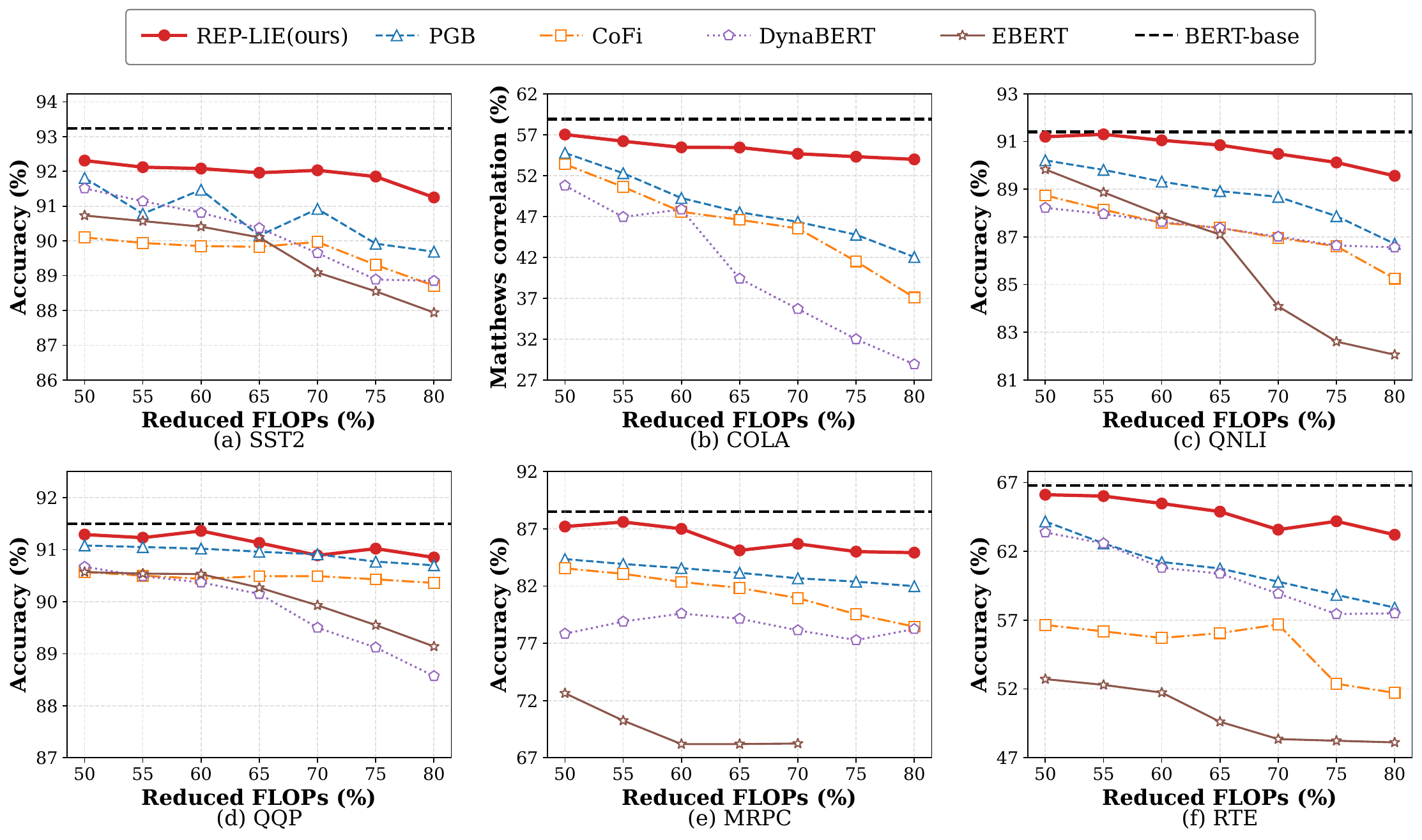}
	\caption{Performance comparison of pruning methods on BERT-{base} under various FLOPs reduction ratios.}
	\label{fig:all_sparsity}
\end{figure*}
We evaluate REP-LIE on two representative model families to validate its generality across scales and architectures.

\textbf{Medium-scale encoder.} We use BERT-base \cite{devlin2019bert}, the standard encoder-only Transformer. We compare against state-of-the-art structured pruning methods: CoFi\cite{CoFiXia2022}, PGB\cite{PGB2025}, DynaBERT\cite{Hou2020DyBERT}, EBERT\cite{LiuLLC2021Ebert}, TinyBERT\cite{Jiao0L20Tiny}, and RECAP\cite{Ilhan2024recap}. The BERT-base model serves as the fine-tuned baseline.

\textbf{Large-scale generative.} We extend evaluation to LLaMA-7B and Mistral-7B, two representative decoder-only LLMs. We compare against LLM-Pruner \cite{ma2023llmpruner}, Compresso\cite{guo2023compresso}, and SlimGPT\cite{slimgpt2024}. The original dense models serve as baselines.

\textbf{Evaluation Metrics}.
For BERT-base, we evaluate on the GLUE benchmark \cite{wang2018glue}. The eight tasks comprise SST-2, MNLI, QQP, QNLI, MRPC, CoLA, STS-B, and RTE. Accuracy serves as the metric for SST-2, MNLI, QNLI, and RTE; Matthews correlation for CoLA; F1 score and accuracy for QQP and MRPC; Pearson and Spearman correlations for STS-B.
For LLaMA-7B and Mistral-7B, we assess language modeling capability and zero-shot commonsense reasoning. Perplexity (PPL) on WikiText-2 validation set (sequence length truncated to 128) serves as the metric for language modeling. PPL measures the model's ability to predict the next token in a sequence, with lower values indicating better language understanding. Zero-shot commonsense reasoning is evaluated on BoolQ, PIQA, HellaSwag, WinoGrande, ARC-easy, ARC-challenge, and OpenBookQA. These tasks assess the model's ability to perform reasoning without task-specific fine-tuning, reflecting its generalization capability in real-world scenarios.

For BERT-base, we follow the standard fine-tuning protocol with learning rate $2 \times 10^{-5}$, batch size 32, and maximum sequence length 128. Data augmentation is excluded from TinyBERT for fair comparison. For LLaMA-7B and Mistral-7B, we apply REP-LIE with LoRA rank $r = 64$ (selected based on the analysis in Section \ref{sec:rank}), scaling factor $\alpha = 16$, and dropout 0.1. All experiments use mixed-precision training (FP16) on NVIDIA vGPU-48GB GPUs.

\subsection{Medium-Scale Model Performance Comparison}\label{sec:50prune}
To ensure a fair comparison, we remove data augmentation from the TinyBERT method. Analyzing Table \ref{tab:50prune}, two key observations can be made. 
\begin{enumerate}	
	\item REP-LIE achieves the best results across all GLUE tasks. Compared to other compression methods, REP-LIE maintains the same parameter count (67M) but suffers minimal model performance loss. Specifically, for the QQP, RTE, and MRPC tasks, the accuracy of our compressed model decreases by only 0.17\%, 0.35\%, and 0.58\%, respectively. EBERT performs the worst among all methods, especially on the MRPC and RTE tasks, with an accuracy of 72.08\% and 52.75\%, respectively, both significantly lower than other methods. This may be due to insufficient retention of model structure or knowledge during EBERT's compression process.
\item Moreover, the wilcoxon signed-rank test (one-tailed) confirms that REP-LIE significantly outperforms all compression methods. The consistent ($p=0.0156$) observed in most comparisons, excluding EBERT, underscores the robustness and effectiveness of REP-LIE method compared to other approaches. While REP-LIE shows no significant performance degradation compared to the full BERT-base model ($p>0.05$).
\end{enumerate}

\subsection{Large-Scale Model Performance Comparison}\label{sec:lager}
To facilitate comparison with prior works\cite{slimgpt2024, ma2023llmpruner, guo2023compresso}, we prune LLaMA-7B at four sparsity ratios (20\%, 25\%, 33\%, and 50\%), yielding models with 5.4B, 5.0B, 4.5B, and 3.4B parameters, respectively. 
Table~\ref{tab:results} presents the PPL and zero-shot commonsense reasoning results of the pruned LLaMA-7B models. The results demonstrate that {REP-LIE} achieves superior performance in both language modeling and reasoning across most sparsity levels and subtasks.

At 20\% sparsity, {REP-LIE} attains the best PPL and the highest average score, closely matching the performance of the original LLaMA-7B model. The advantage of {REP-LIE} becomes more pronounced as the pruning ratio increases. Under 50\% sparsity, it yields an average performance improvement of approximately {5.2\%} over SlimGPT and about {17\%} over LLM-Pruner. Notably, on the BoolQ and ARC-easy datasets, {REP-LIE} shows significant gains; for example, it surpasses SlimGPT by over 5\% on BoolQ and ARC-easy at 50\% sparsity.

We observe that the performance impact of {REP-LIE} is relatively balanced across different tasks. At lower pruning ratios, performance on some tasks, such as BoolQ, PIQA and ARC-e, can even exceed that of the original dense model. 
We attribute 
this improvement to the regularization effect of structured 
pruning: removing redundant attention heads and FFN channels suppresses 
overfitting to fine-tuning datasets, thereby enhancing generalization on 
reasoning-intensive tasks that benefit from cleaner feature extraction.
Furthermore, compared to other methods, the performance degradation of {REP-LIE} is more graceful as sparsity increases, indicating its robustness under aggressive compression. 
In contrast, methods like LLM-Pruner exhibit a steeper decline, particularly on tasks like HellaSwag and ARC-challenge, as their unstable importance estimation inadvertently prunes the deep-layer attention structures and feed-forward pathways critical for multi-hop reasoning.
Moreover, Wilcoxon signed-rank tests further confirm the statistical significance of REP-LIE’s performance gains over baselines on the seven reasoning tasks: most comparisons yield $p \leq 0.0078$, indicating significant superiority of REP-LIE in comparison with large-scale evaluation.
\begin{table*}[htbp]
	\caption{GLUE benchmark results at different high sparsity ratios.\label{tab:highsparsity}}
	\small
	\centering	
	\resizebox{0.75\hsize}{!}{
	\begin{tabular}{llccccccc}
		\toprule
		Sparsity & Method & QNLI & QQP & SST-2 & CoLA & MRPC & RTE & Avg. \\
		\midrule
		0\%  & BERT-base & 91.40 & 91.50 & 93.23 & 58.90 & 88.48 & 66.80 & 81.71 \\
		\midrule
		\multirow{4}{*}{60\%} 
		& CoFi  & 87.86 & 90.56 & 89.95 & 47.76 & 82.85 & 55.91 & 75.81 \\
		& PGB   & 89.53 & 91.12 & 91.57 & 49.55 & 83.87 & 61.43 & 77.84 \\
		& RECAP & 89.01 & 88.72 & 90.95 & 54.35 & 85.44 & 64.99 & 78.91 \\
		& \textbf{REP-LIE(ours)}  & \textbf{91.25} & \textbf{91.34} & \textbf{92.56} & \textbf{56.02} & \textbf{87.00} & \textbf{65.34} & \textbf{80.58} \\
		\midrule
		\multirow{4}{*}{88\%} 
		& CoFi  & 84.70 & 89.80 & 89.00 & 32.10 & 75.30 & 52.40 & 70.55 \\
		& PGB   & 86.40 & 90.10 & 89.60 & 39.50 & 78.20 & 54.30 & 73.01 \\
		& RECAP & 86.54 & 87.23 & 88.21 & 51.30 & 80.10 & 54.24 & 74.60 \\
		& \textbf{REP-LIE(ours)}  & \textbf{89.26} & \textbf{90.65} & \textbf{91.20} & \textbf{53.21} & \textbf{84.52} & \textbf{63.01} & \textbf{78.64} \\
		\bottomrule
	\end{tabular}}
\end{table*}
\begin{table*}[htbp]
	\small
	\centering
	\caption{Comparison of model performance and  GPU memory footprint(MB) for BERT-base between REP-LIE and other compression strategies.}
	\label{tab:nlu_performance}
	\resizebox{0.80\hsize}{!}{
	\begin{tabular}{clccccccc}
		\toprule
	{Sparsity}&	{Method} & {CoLA } & {SST-2} & {MRPC } & {QQP} & {QNLI} & {RTE} & \makecell{Memory\\(MB)} \\
		\midrule
		0\% & BERT-base   & 58.90 & 93.20 & 88.50 & 90.89 & 91.40 & 66.80 & 1553 \\
		\midrule
		\multirow{5}{*}{50\%}
	&	Head-FT     & 51.10 & 89.75 & 85.44 & 88.05 & 90.66 & 64.99 & 524 \\
	&	Pre-FT      & 53.17 & 91.16 & 83.88 & 89.47 & 90.12 & 62.45 & 957 \\
	&	Post-FT     & \textbf{57.78} & \textbf{92.70} & \textbf{88.20} & 88.52 & 90.89 & \textbf{66.89} & 1553 \\
	&	RECAP       & 56.56 & 92.20 & 86.56 & 89.39 & 90.74 & 65.98 & 702 \\
	&	\textbf{REP-LIE(ours)} & \textbf{57.23} & \textbf{92.33} & \textbf{87.90} & \textbf{90.20} & \textbf{90.96} & 66.45 & \textbf{476} \\
		\bottomrule
	\end{tabular}}
\end{table*}

\subsection{Medium Model Evaluation at High Compression Ratio}\label{sec:mehigh}
The performance of REP-LIE is further evaluated under the condition of high sparsity levels, which is shown in Table \ref{tab:highsparsity}. And Fig.~\ref{fig:all_sparsity} shows the performance comparisons for REP-LIE and other works under the condition of varying FLOPs reduction ratios. In summary, results in Table \ref{tab:highsparsity} and Fig.~\ref{fig:all_sparsity} underscore two distinct advantages of REP-LIE:
\begin{enumerate}
	\item Robust multi-task performance under varying compression levels. REP-LIE maintains stable performance across GLUE tasks at both different high sparsity ratios and varying FLOPs reduction ratios, significantly outperforming mainstream compression methods. For example, when the sparsity rate reaches 88\%, the average performance remains at 78.64\%, far exceeding other methods. Notably, on the MRPC task, the performance drops by only 3.96\% compared with the baseline, significantly outperforming methods like CoFi and PGB.
	\item Superior stability via progressive pruning. In contrast to one-shot pruning methods such as EBERT and CoFi suffering from performance degradation under high sparsity, REP-LIE employs a progressive pruning strategy and optimal mask search to better retain model performance, as depicted in Section \ref{progressive pruning}. On SST-2 and QQP, REP-LIE retains performance close to the baseline, with negligible degradation across pruning levels. For the tasks of QNLI and MRPC, EBERT and CoFi show significant performance degradation. The experimental results further validate the stability of REP-LIE.
\end{enumerate}

\subsection{Memory Usage Comparison}\label{sec:memory}
To evaluate the memory efficiency of REP-LIE, we report the corresponding memory footprint and compare it with that of state-of-the-art works. In Head-FT, we freeze the pre-trained model weights and only update the parameters of the injected task head to reduce the GPU memory footprint. In Pre-FT (pre-finetune pruning), we finetune the model after performing the pruning operation. In Post-FT (post-finetune pruning), we finetune the model before and after pruning, following \cite{molchanov2019importance}. RECAP is also compared, which prunes the model using Taylor-approximation-based importance estimation at each iteration and then updates only a subset of the pruned model weights according to the Fisher-information criterion. For all pruning techniques, we report the result for sparsity = 50\%. The experimental results are shown in Table \ref{tab:nlu_performance}.

It can be seen that REP-LIE achieves the lowest memory footprint along with competitive model performance. Specifically, compared to Head-FT, Pre-FT and RECAP, REP-LIE achieves the best outcomes in both model performance and memory footprint. And REP-LIE only requires 476MB of memory, showing competitive model performance and reducing memory consumption by 69.4\% compared to Post-FT. 
Empirically, the enhanced memory utilization efficiency of REP-LIE can be primarily ascribed to two key design:
\begin{enumerate}
\item REP-LIE utilizes low-rank gradients to obviate the need for computing gradients of pre-trained weights, thereby significantly reducing the computational cost and memory footprint associated with importance estimation. 
\item Rather than performing full-model finetuning, REP-LIE selectively updates low-rank matrices in LoRA to recover model performance. This strategy effectively mitigates the substantial memory footprint associated with comprehensive parameter updates, thereby enhancing memory efficiency. 
\end{enumerate}
\begin{table*}[!t]
	\centering
	\scriptsize
	\caption{Ablation study on pruning criterion for BERT-base.}
	\label{tab:criterion}
	\setlength{\tabcolsep}{3pt}
	\resizebox{0.95\hsize}{!}{
		\begin{tabular}{@{}>{\raggedright\arraybackslash}p{2.2cm} 
				*{8}{>{\centering\arraybackslash}p{0.7cm}}
				>{\centering\arraybackslash}p{0.7cm}
				>{\centering\arraybackslash}p{0.7cm}
				>{\centering\arraybackslash}p{0.7cm}
				>{\centering\arraybackslash}p{0.7cm}@{}}
			\toprule
			Pruning criterion & QQP & QNLI & SST-2 & CoLA & MRPC & RTE & MNLI & STSB & Avg. & \makecell{Time\\(s)} & \makecell{Mem.\\(MB)} & \makecell{FLOPs\\(G)} \\
			\midrule
			Base model (dense) & 91.50 & 91.40 & 93.20 & 58.90 & 88.50 & 66.80 & 84.60 & 88.90 & 82.97 & -- & 1553 & -- \\
			\midrule
			+ Weight magnitude & 90.71 & 89.10 & 92.00 & 54.40 & 86.20 & 65.63 & 83.12 & 84.60 & 80.72 & 115.6 & 476 & \textbf{0.00} \\
			+ Gradient-based & 91.12 & \textbf{90.71} & \textbf{92.50} & \textbf{57.43} & 86.96 & \textbf{66.50} & 83.85 & 86.50 & 81.87 & 180.8 & 810 & 0.26 \\
			+ LoRA-based (Ours) & \textbf{91.33} & 90.96 & 92.33 & 57.23 & \textbf{87.90} & 66.45 & \textbf{83.98} & \textbf{87.01} & \textbf{82.14} & 120.4 & \textbf{476} & 0.01 \\
			\bottomrule
	\end{tabular}}
\end{table*}
\begin{table*}[t]
	\centering
	\scriptsize
	\caption{Ablation study on pruning criterion for LLaMA-7B. The PPL is evaluated on Wikitext2 and the Avg. is computed across seven Commonsense Reasoning datasets. }
	\label{tab:1}
	\setlength{\tabcolsep}{2pt}
	\resizebox{0.95\hsize}{!}{
		\begin{tabular}{@{}>{\raggedright\arraybackslash}p{2.2cm} 
				*{8}{>{\centering\arraybackslash}p{0.7cm}}
				>{\centering\arraybackslash}p{0.7cm}
				>{\centering\arraybackslash}p{0.7cm}
				>{\centering\arraybackslash}p{0.7cm}
				>{\centering\arraybackslash}p{0.7cm}@{}}
			\toprule
			Pruning criterion & PPL$\downarrow$ & BoolQ & PIQA & HellaS & WinoG & ARC-e & ARC-c & OBQA & Avg. & \makecell{Time\\(s)} & \makecell{Mem.\\(MB)} & \makecell{FLOPs\\(G)} \\
			\midrule
			Base model (dense) & 12.63 & 75.08 & 79.16 & 76.20 & 70.00 & 72.89 & 44.88 & 44.40 & 66.09 & -- & 12884 & -- \\
			\midrule
			+ Weight magnitude & 32.59 & 60.21 & 68.88 & 58.25 & 57.62 & 56.56 & 29.73 & 38.20 & 52.79 & \textbf{982.3} & 5459 & \textbf{0.00} \\
			+ Gradient-based & 30.45 & 69.88 & 72.53 & 59.83 & 60.23 & \textbf{58.89} & 31.62 & \textbf{40.98} & 56.28 & 1948.7 & 11601 & 13.40 \\
			+ LoRA-based (Ours) & \textbf{28.46} & \textbf{70.58} & \textbf{73.26} & \textbf{60.23} & \textbf{60.98} & 58.52 & \textbf{32.40} & 40.02 & \textbf{56.57} & 983.6 & \textbf{5459} & 0.65 \\
			\bottomrule
	\end{tabular}}
\end{table*}
\begin{table*}[t]
	\centering
	\scriptsize
	\caption{Ablation study on pruning criterion for Mistral-7B. The PPL is evaluated on Wikitext2 and the Avg. is computed across seven Commonsense Reasoning datasets.}
	\label{tab:2}
	\setlength{\tabcolsep}{2pt}
	\resizebox{0.95\hsize}{!}{
		\begin{tabular}{@{}>{\raggedright\arraybackslash}p{2.2cm} 
				*{8}{>{\centering\arraybackslash}p{0.7cm}}
				>{\centering\arraybackslash}p{0.7cm}
				>{\centering\arraybackslash}p{0.7cm}
				>{\centering\arraybackslash}p{0.7cm}
				>{\centering\arraybackslash}p{0.7cm}@{}}
			\toprule
			Pruning criterion & PPL$\downarrow$ & BoolQ & PIQA & HellaS & WinoG & ARC-e & ARC-c & OBQA & Avg. & \makecell{Time\\(s)} & \makecell{Mem.\\(MB)} & \makecell{FLOPs\\(G)} \\
			\midrule
			Base model (dense) & 5.25 & 80.94 & 82.80 & 81.30 & 75.30 & 80.00 & 55.50 & 45.40 & 71.60 & -- & 11263 & -- \\
			\midrule
			+ Weight magnitude & 26.64 & 63.45 & 70.76 & 61.26 & 60.87 & 68.52 & 35.26 & 35.29 & 56.49 & \textbf{986.8} & 5392 & \textbf{0.00} \\
			+ Gradient-based & 18.12 & 74.23 & 77.53 & \textbf{74.86} & 68.01 & 73.13 & \textbf{41.62} & 38.98 & 64.05 & 2089.6 & 10589 & 14.48 \\
			+ LoRA-based (Ours) & \textbf{15.36} & \textbf{75.11} & \textbf{77.60} & 74.01 & \textbf{69.27} & \textbf{73.37} & 40.83 & \textbf{39.20} & \textbf{64.19} & 996.9 & \textbf{5392} & 0.70 \\
			\bottomrule
	\end{tabular}}
\end{table*}

These two mechanisms work in tandem to enable REP-LIE to achieve a memory footprint of only 476MB, representing a 69.4\% reduction compared to Post-FT, while maintaining competitive performance across various natural language understanding tasks.
\begin{table}[!t]
	\centering
	\caption{Pruning runtime and peak GPU memory usage of REP-LIE and two advanced methods on LLaMA-7B.}
	\label{tab:pruning_runtime_com}
	\resizebox{0.85\hsize}{!}{
		\begin{tabular}{lcccccc}
			\toprule
			{Method} & 
            \makecell{Runtime\\@33\%sparsity} & \makecell{Runtime\\@50\%sparsity} & \makecell{Memory\\(MB)} \\
			\midrule
			LLM-Pruner  & 895.4s  & 1290.8s & 6890 \\
			SlimGPT  & 783.0s  & 1074.0s & 7375 \\
			  \textbf{REP-LIE (ours)} & \textbf{594.2s} & \textbf{983.6s} & \textbf{5459} \\
			\bottomrule
	\end{tabular}}
\end{table}
\begin{table}[!t]
	\centering
	\caption{Pruning runtime and peak GPU memory usage of REP-LIE at different sparsity ratios on different models.}
	\label{tab:pruning_runtime}
	\resizebox{0.85\hsize}{!}{
		\begin{tabular}{ccccc}
			\toprule
			\makecell{Model} & \makecell{Runtime\\@33\%sparsity} & \makecell{Runtime\\@50\%sparsity} & \makecell{Memory\\(MB)} \\
			\midrule
			BERT-base  & 86.2s  & 120.4s & 476 \\
			LLaMA-7B  & 594.2s  & 983.6s & 5459 \\
			Mistral-7B & 598.6s & 996.9s & 5392 \\
			\bottomrule
	\end{tabular}}
\end{table}
\begin{table}[!t]
	\centering
	\footnotesize
	\caption{Inference efficiency of pruned LLaMA-7B at different sparsity ratios.}
	\label{tab:inference_latency}
	\setlength{\tabcolsep}{2pt}
	\resizebox{0.99\hsize}{!}{
		\begin{tabular}{@{}>{\centering\arraybackslash}p{1.2cm} 
				>{\centering\arraybackslash}p{1.2cm}
				>{\centering\arraybackslash}p{1.2cm}
				>{\centering\arraybackslash}p{1.4cm}
				>{\centering\arraybackslash}p{1.2cm}
				>{\centering\arraybackslash}p{1.2cm}@{}}
			\toprule
			\makecell{Sparsity} & \makecell{\#Params} & \makecell{Latency\\(ms)} & \makecell{Throughput\\(token/s)} & \makecell{Memory\\(MB)} & \makecell{FLOPs\\(T)} \\
			\midrule
			0\%  & 6.7B & 14.87 & 8608 & 28796 & 1.7 \\
			33\%        & 5.4B & 12.57 & 10183 & 24697 & 1.4 \\
			50\%        & 3.4B & 8.87  & 14430 & 13587 & 0.9 \\
			\bottomrule
	\end{tabular}}
\end{table}

\subsection{Efficiency Evaluation on Medium and Large Model}\label{sec:efficiency}
We further compare the pruning efficiency of REP-LIE against advanced baseline methods. As shown in Table~\ref{tab:pruning_runtime_com}, REP-LIE achieves the fastest pruning time and lowest memory footprint across all sparsity levels. This efficiency advantage mainly arises from the low-rank gradient-based importance estimation, which avoids full-gradient computation by leveraging only LoRA's compact matrices, coupled with lightweight LoRA finetuning that eliminates costly full-parameter updates. To provide a more comprehensive view of REP-LIE's efficiency, Tables~\ref{tab:pruning_runtime} and~\ref{tab:inference_latency} report its pruning overhead and inference speedup across different model scales, respectively.

\textbf{Pruning overhead}. Pruning 50\% of the parameters on LLaMA-7B consumes 5.4 GB of GPU memory and completes in 16 minutes, which is 57.6\% of the memory required to load the full model. This is because importance
estimation operates exclusively on the low-rank LoRA matrices rather than the
entire gradient. Mistral-7B achieves comparable memory usage to
LLaMA-7B, which is attributable to its Grouped-Query Attention(GQA) mechanism, which reduces the number
of KV heads from 32 to 8 and thus lowers the memory footprint of attention
computations during fine-tuning. For smaller models such as BERT-base, the
same procedure completes in approximately 2 minutes, demonstrating that the
overhead scales efficiently with model size.
	
\textbf{Inference speedup.} The pruned models yield substantial deployment gains.
At 50\% sparsity, the pruned LLaMA-7B achieves a 47\% reduction in peak GPU memory consumption, a 40\% reduction in latency, and a 68\% increase in throughput, relative to the dense model. The inference FLOPs
are also nearly halved. These results confirm that REP-LIE
directly translates parameter reduction into tangible inference improvements, suiting resource-constrained deployment.
\begin{figure*}[ht]
	\centering
	\includegraphics[scale=0.45]{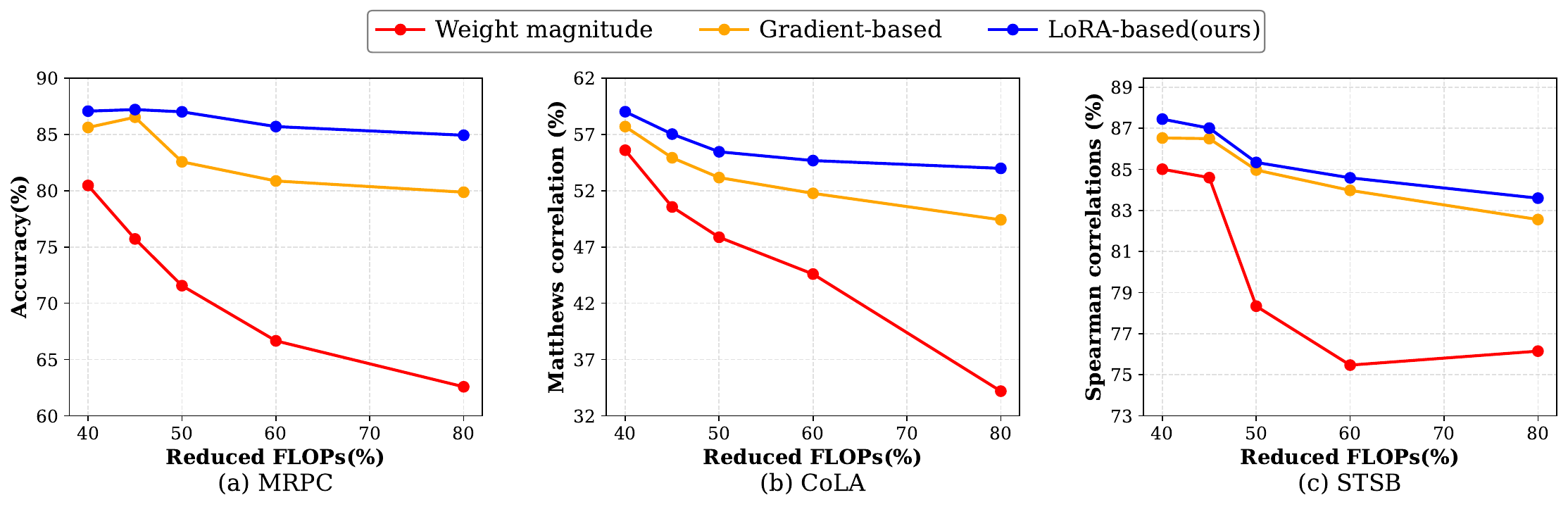} 
	\caption{Performance comparison of three pruning criteria on BERT-base under various FLOPs reduction ratios.}
	\label{fig:criterion}
\end{figure*}
\begin{figure*}[ht]
	\centering
	\includegraphics[scale=0.45]{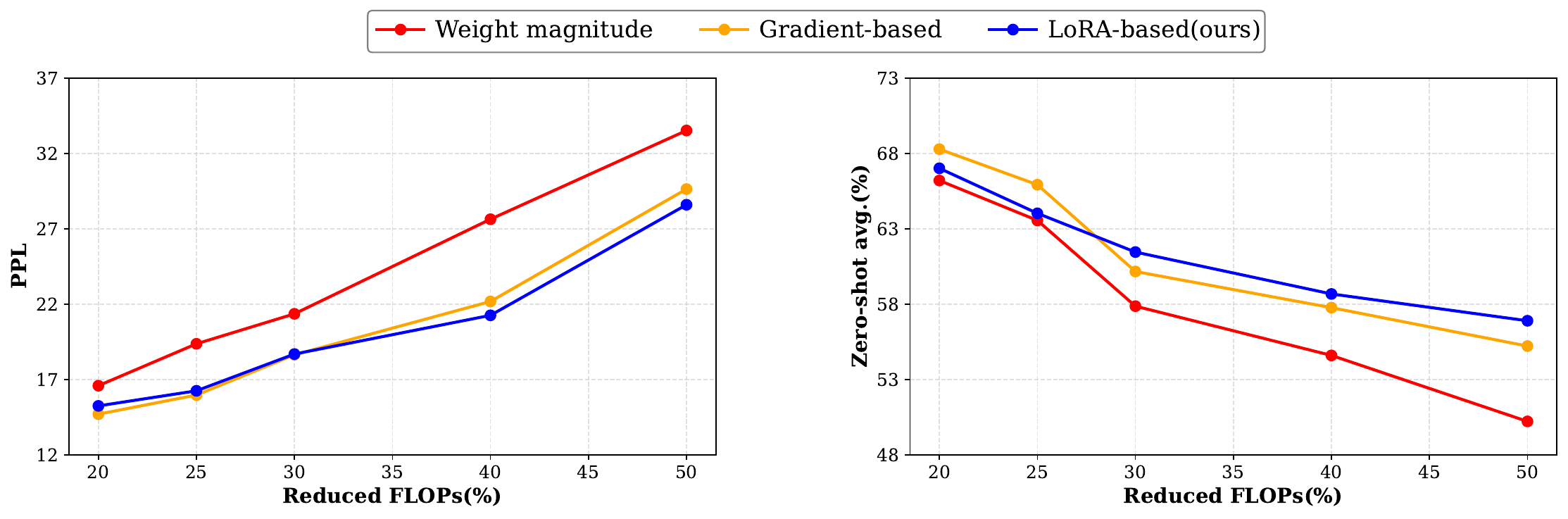} 
	\caption{PPL \text{\&} Commonsense Reasoning zero-shot performance comparison of three pruning criteria on LLaMA-7B under various FLOPs reduction ratios.}
	\label{fig:pruning_comparison_llama}
\end{figure*}
\begin{figure*}[!t]
	\centering
	\includegraphics[scale=0.45]{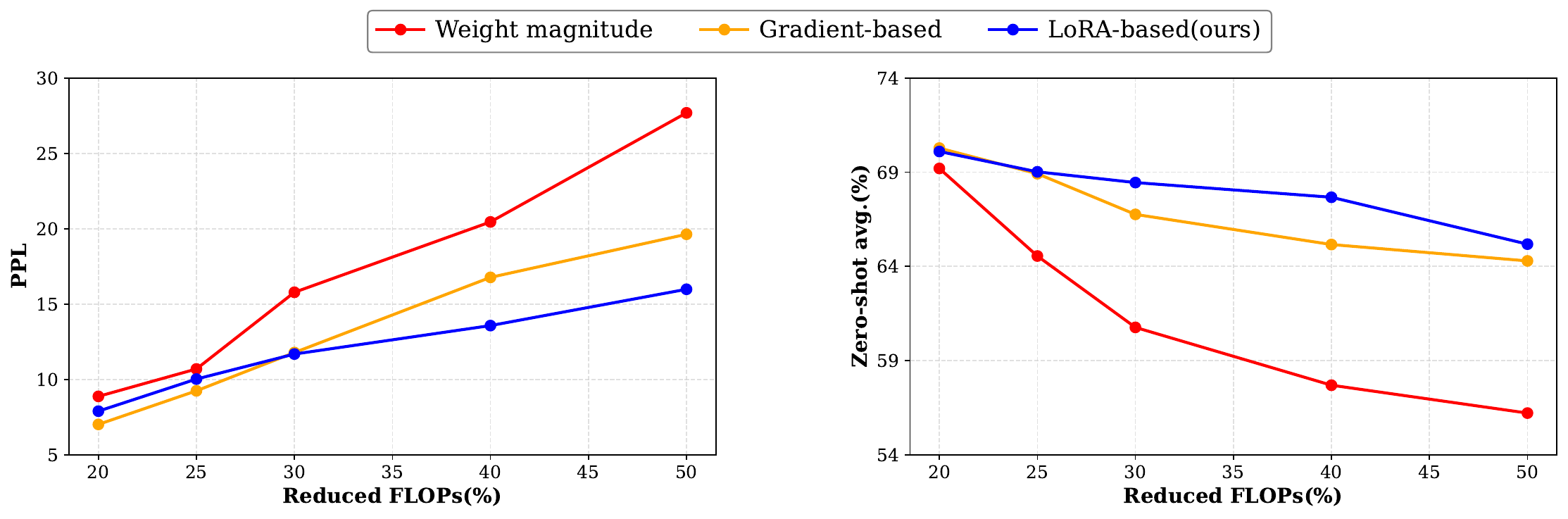} 
	\caption{PPL \text{\&} Commonsense Reasoning zero-shot performance comparison of three pruning criteria on Mistral-7B under various FLOPs reduction ratios.}
	\label{fig:pruning_comparison_mistral}
\end{figure*}
\begin{figure*}[htbp] 
	\centering
	\includegraphics[scale=0.4]{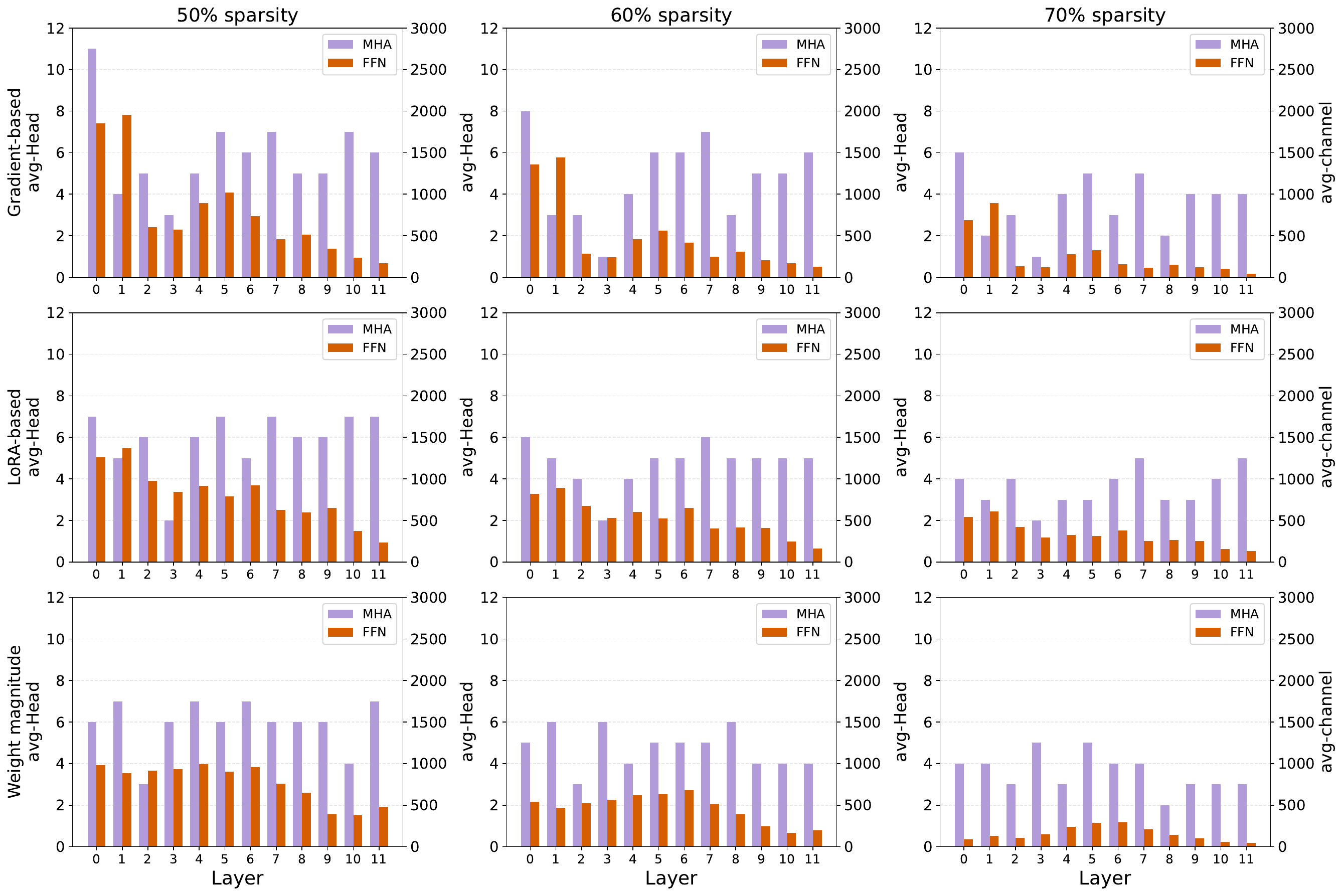} 
	\caption{The visualization of the average number of unpruned attention heads in MHA layers and unpruned channels in FFN layers under different sparsities.}
	\label{fig:double}
\end{figure*}

\begin{figure}[htbp]
	\includegraphics[scale=0.44]{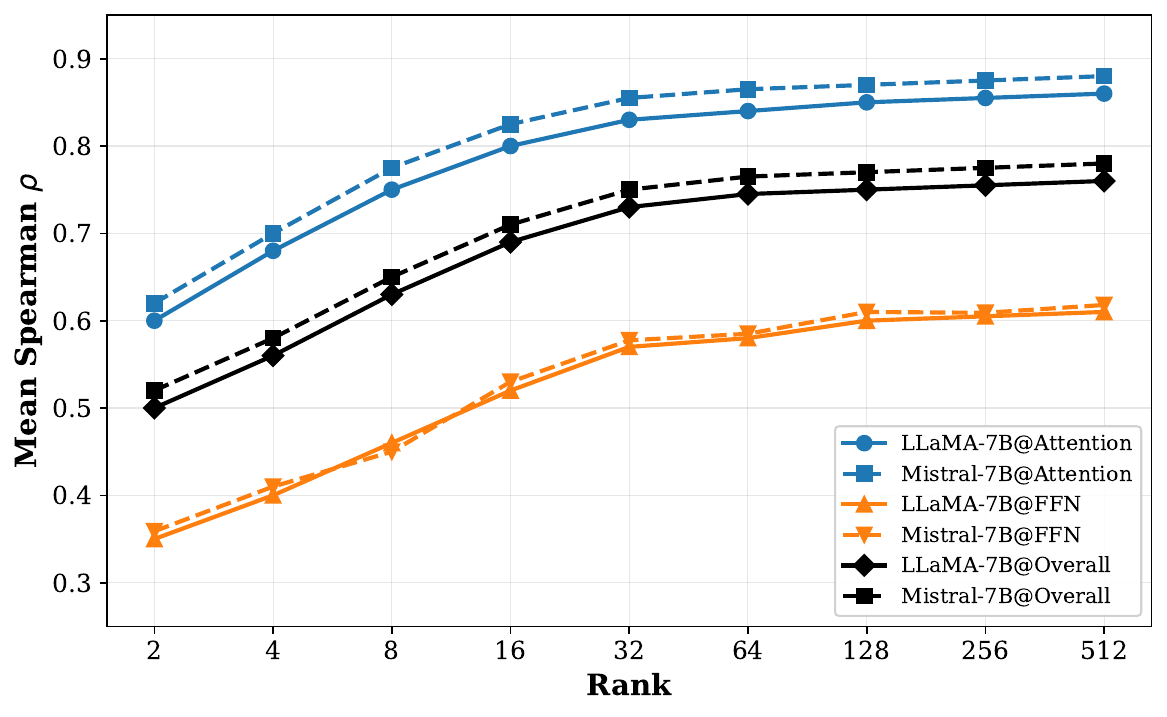}
	\caption{The mean spearman correlation between low-rank and full weight gradients across different ranks and models.}
	\label{fig:rank_corr}
\end{figure}

\subsection{Pruning Criterion Analysis \label{sec:rank}}\label{sec:criterion}
\textbf{Performance and Efficiency Comparison.}
We further evaluate three importance estimation strategies (weight magnitude, gradient-based, and LoRA-based) across BERT-base, LLaMA-7B, and Mistral-7B, at a sparsity of 50\%,  as reported in Tables~\ref{tab:criterion}, \ref{tab:1}, and~\ref{tab:2}.

Two consistent patterns emerge across all models. First, gradient-based and
LoRA-based achieve comparable task performance, while weight magnitude lags
behind in all cases, confirming that gradient information is essential for
effective structural importance estimation. Second, LoRA-based achieves this
comparable performance at a fraction of the computational cost of gradient-based:
GPU memory is reduced by 49\%--50\% and
FLOPs for importance estimation by over 96\%,
while weight magnitude avoids additional FLOPs but at a severe accuracy cost.
Beyond these shared patterns, an architecture-dependent effect appears on
Mistral-7B: the performance gap between LoRA-based and weight magnitude is
amplified compared to LLaMA-7B. LoRA-based retains 89.7\% of Mistral's
original zero-shot performance versus 78.9\% for weight magnitude. This amplification
demonstrates that the advantage of our method grows with the
architectural sophistication of the target model.

\textbf{Robustness Evaluation Across Compression Ratios.}
To further investigate the robustness of different pruning criteria across
varying compression levels, we conduct experiments on BERT-base, LLaMA-7B,
and Mistral-7B under increasing FLOPs reduction ratios, as depicted in
Figs. \ref{fig:criterion}, \ref{fig:pruning_comparison_llama}, and
\ref{fig:pruning_comparison_mistral}.

Across all three models, the three criteria perform comparably at low
compression ratios but diverge under aggressive pruning. The performance advantage of the LoRA-based method widens with an increasing FLOPs reduction ratio.

On BERT-base (Fig. \ref{fig:criterion}), the accuracy gap between criteria
widens most sharply at 80\% FLOPs reduction, reaching about 15.7\% on CoLA
and 8.2\% on MRPC, indicating that gradient-aware criteria are
particularly valuable when compression is severe on encoder models.

On LLaMA-7B (Fig. \ref{fig:pruning_comparison_llama}), a notable asymmetry
emerges under weight magnitude pruning: its PPL degradation is relatively
moderate compared to the steeper drop in zero-shot reasoning performance,
suggesting that magnitude-based removal inadvertently spares parameters
critical for perplexity while pruning those essential for higher-level
reasoning tasks.

On Mistral-7B (Fig. \ref{fig:pruning_comparison_mistral}), the advantage of
LoRA-based over weight magnitude is amplified relative to LLaMA-7B, owing to
Mistral's architectural design: SwiGLU's gated FFN introduces more complex
parameter interactions that gradient-aware methods capture more faithfully
than magnitude alone, while GQA concentrates gradient signals across fewer
KV heads, enhancing the discriminability of low-rank importance estimates.

\textbf{Mechanistic Analysis.}
To explore how various pruning criteria impact the internal structure of Transformer models, we visualize the number of unpruned attention heads in MHA and unpruned channels in FFN across all layers under varying sparsity levels, as shown in Fig.~\ref{fig:double}. The observations are concluded as follows:
\begin{enumerate}
\item Across all methods, the retention ratio of channels in the FFN layer is significantly lower than that of attention heads in the MHA layer, indicating that these methods generally prefer to retain the attention mechanism to maintain the model performance.
\item Beginning with the \textit{9-th} layer, the number of FFN channels decreases significantly, particularly at higher sparsity levels (60\% and 70\%). This indicates that deeper layers exhibit greater redundancy and are thus more suitable for compressing.

\item Attention heads in the first few layers (such as layers 0 and 1) are hardly pruned, which is more pronounced in the LoRA method, reflecting the importance of shallow-layer attention in maintaining the input representation ability.
\end{enumerate}

We have designed a systematic set of experiments to evaluate the approximation quality of low-rank gradients relative to full-weight gradients for importance estimation. 
Fig.~\ref{fig:rank_corr} presents the mean spearman rank correlation $\rho$ between importance scores derived from low-rank gradients and those from full-weight gradients, evaluated
across LoRA ranks $r \in \{2, 4, 8, 16, 32, 64, 128, 256, 512\}$ on both LLaMA-7B
and Mistral-7B, reported separately for attention layers, FFN layers, and all
layers combined. The key observations are summarized as follows:
\begin{enumerate}
	\item {Low-rank gradients preserve dominant importance structure even at minimal ranks.} Even at $r = 2$, the overall correlation exceeds 0.5, a threshold widely 
	recognized as denoting strong correlation in \cite{cohen1988statistical}, indicating that the coarsest approximation captures the dominant ranking structure. The correlation improves rapidly as rank increases, reaching $\rho \approx 0.72$ for all layers and $\rho \approx 0.83$ for Attention layers at $r = 32$. The curve flattens markedly beyond $r = 64$, with marginal gains diminishing to $\Delta \rho < 0.02$ when doubling to 128.  Thus, $r = 64$ achieves high-fidelity approximation ($\rho > 0.75$ overall, $\rho > 0.85$ for Attention) with negligible degradation relative to full-rank estimation, while maintaining computational efficiency. Ranks below 32 remain viable for resource-constrained scenarios but incur measurable approximation error, particularly for FFN layers.
	
	\item Attention layers exhibit superior low-rank approximability compared to FFN layers. This disparity can be attributed to the intrinsic structural and functional properties of these components. The attention weight matrices (e.g., $W_q, W_k, W_v$) encode concentrated semantic and positional information, their gradients likely possess a lower intrinsic rank, or their principal components are more readily captured by low-rank subspaces. In contrast, FFN layers, especially those that employ advanced activation functions like SwiGLU, perform more complex nonlinear feature transformations, resulting in higher-dimensional and more dispersed parameter spaces. This produces a flatter gradient spectrum, necessitating higher ranks to achieve comparable approximation fidelity.
	
	\item Mistral-7B consistently achieves higher correlation than LLaMA-7B. This
	is attributable to Mistral-7B's GQA, lowering the effective
	dimensionality of attention gradients and making them more amenable to
	low-rank approximation.
\end{enumerate}

\section{Conclusion}\label{sec:conclu}
This paper proposes REP-LIE, a structured pruning framework for efficient
compression and lightweight finetuning of large-scale Transformer-based language
models. REP-LIE evaluates weight importance using the gradients of LoRA low-rank
matrices, eliminating the need for full-gradient computation and substantially
reducing memory and computational overhead during pruning. A stability score
mechanism further ensures reliable importance estimation, and a lightweight
finetuning strategy recovers model performance with minimal resource consumption.
Comprehensive experiments on both encoder models and large-scale decoder-only
LLMs demonstrate that REP-LIE achieves state-of-the-art performance across
diverse architectures.
By unifying importance estimation and performance recovery within
the same low-rank parameter space, REP-LIE shifts the prevailing paradigm from
sequential pruning-then-finetuning to simultaneous pruning-and-finetuning. This suggests a fundamental shift in improving the resource efficiency of deployment pipelines, moving from the isolated optimization of individual stages toward a co-designed integration of compression and adaptation as interdependent processes. 

\textbf{Future work.} Although REP-LIE demonstrates the effectiveness of low-rank importance estimation in standard pretraining-finetuning paradigms, its generalizability to diverse learning scenarios and potential for compositional compression remain underexplored. That is, REP-LIE establishes a foundation for structured pruning of standard pretrained Transformers. Building upon this foundation, an important extension would be to investigate its effectiveness on already-compressed models, such as knowledge-distilled variants or quantized models. These compact models present additional challenges due to their shallower architectures, more aggressive compression, and potentially altered gradient structures, which require the developed stable, rank-aware pruning mechanism of REP-LIE as a prerequisite. Joint optimization of pruning and quantization is another promising avenue, as the resource efficiency gained from structural compression can be further amplified when combined with reduced-precision inference.

\bibliographystyle{IEEEtran} 

\bibliography{refs} 

\vfill

\end{document}